\documentclass{article}
\pdfoutput=1

     \usepackage[nonatbib,preprint]{neurips_2019}

\usepackage[utf8]{inputenc} 
\usepackage[T1]{fontenc}    
\usepackage{hyperref}       
\usepackage{url}            
\usepackage{booktabs}       
\usepackage{amsfonts}       
\usepackage{nicefrac}       
\usepackage{microtype}      
\usepackage{amsmath}
\usepackage{amssymb}
\usepackage{amsthm}
\usepackage{subfigure}
\usepackage{graphicx}

\title{Contrastive Learning for Lifted Networks}

\author{%
  Christopher Zach 
  \\
  Chalmers University of Technology\\
  Gothenburg, Sweden \\
  \texttt{christopher.m.zach@gmail.com} \\
  \And
  Virginia Estellers \\
  Microsoft \\
  Cambridge, UK \\
  \texttt{virginia.estellers@gmail.com}
}

\theoremstyle{plain}

\theoremstyle{definition}
\newtheorem{remark}{Remark}

\providecommand{\norm}[1]{\lVert#1\rVert}
\providecommand{\Norm}[1]{\left\lVert#1\right\rVert}

\newcommand{\eps}{\varepsilon}

\def\m#1{\ensuremath{\mathtt{#1}}}
\def\v#1{\ensuremath{\mathbf{#1}}}
\def\mI{\m I}

\def\mW{\m W}

\def\sC{\ensuremath{\mathcal{C}}}

\begin{document}

\maketitle

\begin{abstract}
  In this work we address supervised learning of neural networks via lifted network
  formulations. Lifted networks are interesting because they allow training on
  massively parallel hardware and assign energy models to discriminatively
  trained neural networks. We demonstrate that the training methods for lifted
  networks proposed in the literature have significant limitations and show how 
  to use a contrastive loss to address those limitations. We
  demonstrate that this contrastive training approximates back-propagation in theory
  and in practice and that it is superior to the training objective regularly used
  for lifted networks.
\end{abstract}

\section{Introduction}
\label{sec:intro}

Almost all methods of supervised training of deep neural networks (DNNs) rely on back-propagation (in combination with stochastic gradient
descent or an accelerated version thereof) to adjust the network parameters in
order to minimize a given loss function. While back-propagation is highly
successful in practice, it has some limitations: (i) it can suffer from the
vanishing and exploding gradient problem, (ii) its implicit use of
fine-grained synchronization limits its implementation on massively
parallel hardware, and (iii) there is evidence that back-propagation is
not the basis of learning in biological systems.

\emph{Lifted networks} have been proposed mainly to utilize massively parallel
hardware to train deep neural
networks~\cite{carreira2014distributed,zhang2016efficient,zhang2017convergent,taylor2016training,lau2018proximal}. Lifted
networks introduce explicit variables to represent the activations of network
units, and these activations are determined implicitly as minimizers of an
underlying optimization problem (which we call the \emph{network energy}). In
strictly layered DNN architectures the learning problem decouples into
layer-wise subproblems over the weights once the activations are fixed (and vice versa). This
enables massively parallel optimization of the decoupled sub-problems. We are
especially interested in the setting when the network energy is strictly
convex because strictly convex energy have a single global minimum and can be optimized with parallel updates of
the activations with (block) coordinate descent (e.g.~\cite{sontag2009tree}).
This is not a limitation because strictly convex energies approximate the computations in feed-forward networks with arbitrary precision for a range of
non-linearities. 

Training of lifted networks is conducted by augmenting the network energy with
a loss term that steers the network output to the correct label. This objective is minimized with
respect to the network weights and the activations (for a whole training set),
i.e.\ the optimal weights lead to the smallest (average) \emph{loss-augmented}
network energies over the training set. At test time
the network's output is inferred from the original, non-augmented network
energy. Hence, there is a mismatch on how network activations and outputs are
determined during the training and testing phases, which leads to inferior prediction performance. We demonstrate that
training lifted networks solely with loss-augmented network energies largely
keeps the network in the linear regime and does therefore not leverage the
expressive power of non-linear units.

In this work we propose to utilize a \emph{contrastive} training objective to
address the shortcomings of standard methods to train lifted
networks. Essentially, we reconcile the training and inference phase of lifted
networks by applying ideas of contrastive Hebbian
learning~\cite{movellan1991contrastive,xie2003equivalence} to the lifted
formulation. To this purpose, instead of determining the networks parameters
that minimize the loss-augmented energy over the training samples, we find the
network parameters that minimize a contrastive objective, which ensures that
minima of the network energy agree with minima of the loss-augmented
energy used to train the lifted network. Our contrastive loss generalizes to
any convex loss function and substantially improves the performance of lifted
networks.
We will show that training lifted networks with a contrastive approach
approximates back-propagation. Hence, we connect energy-based models and
back-propagation based discriminative training.

Compared to traditional contrastive Hebbian
learning~\cite{movellan1991contrastive,xie2003equivalence,scellier2017equilibrium},
the convex formulation of lifted networks speeds up the computation of the
contrastive loss and allows for distributed training of deep neural
networks. The connection between back-propagation and energy-based models
might offer a better understanding of energy landscapes and generalization
ability of different network parameters.

\paragraph{Related Work}
Energy-based models in machine learning have a long history, with Hopfield
nets~\cite{hopfield1982neural}, Boltzmann machines~\cite{ackley1985learning}
and restricted Boltzmann machines~\cite{smolensky1986,hinton2002training} as
prominent examples. Contrastive learning methods using non-convex network
energies, which relate to variants of Hebbian learning, are proposed
in~\cite{movellan1991contrastive,xie2003equivalence,scellier2017equilibrium}. Among this works, \cite{xie2003equivalence} is noteworthy because it
proves the connection between back-propagation and contrastive loss-based
optimization using a particular non-convex network energy.

Several recent works also formulate training of lifted networks in a lifted space for parallel computation. \cite{carreira2014distributed} proposes a quadratic relaxation for the
computations in a feedforward network, and thereby introduces ``auxiliary
coordinates'', i.e.\ network activations as explicit variables, to obtain a
highly distributed learning algorithm. Further algorithmic improvements based
on this or similar quadratic relaxation were proposed
in~\cite{taylor2016training,lau2018proximal,gotmare2018decoupling}.
It was recognized in~\cite{zhang2017convergent,askari2018lifted}, that ReLU
and other non-linearities can be approximated using quadratic network energies
with bounds constraints on the activations. \cite{gu2018fenchel,li2018lifted}
extend this construction to larger classes of non-linearities and obtain
block multi-convex network energies. In particular, \cite{li2018lifted}
proposes non-convex energies, where inference of activation by minimization
yields exactly the feedforward pass in standard DNNs.

Other methods aiming to replace or to modify back-propagation are difference
target propagation~\cite{lee2015difference}, proximal
back-propagation~\cite{frerix2017proximal}, and approaches based on activation
statistics~\cite{choromanska2018beyond} or DC
programming~\cite{berrada2016trusting}. Synthetic gradients aim to avoid the
need of synchronization in back-propagation~\cite{jaderberg2017decoupled}.

\section{Lifted Networks}
\label{sec:contrastive_loss}

This section reviews lifted networks and introduces some of the notation used
in subsequent sections. Lifted networks determine the internal network
activations implicitly by solving an optimization problem instead of relying
on explicitly provided mappings, e.g.\ in feed-forward networks. By solving an
optimization problem, lifted networks usually also allow activations in later
layers to influence earlier layers.

\paragraph{Notations}
For a convex set $\sC$ we use $\imath_\sC$ to denote the indicator
function, $\imath_\sC(x) = 0$ iff $x\in \sC$ and
$\imath_\sC(x)=\infty$ otherwise. We write $\Pi_\sC(x)$ for the projection of
$x$ into a convex set $\sC$.

\paragraph{Lifted networks}
Lifted networks were initially introduced to enable massively parallel
implementations of deep neural networks~\cite{carreira2014distributed} (with
later extensions
of~\cite{taylor2016training,frerix2017proximal,zhang2017convergent,lau2018proximal}).
The core idea of using convex energies to determine hidden unit activations is
by observing that ReLU non-linearities (and also other ones such as
hard-sigmoid and leaky ReLU) can be stated as proximal
operators~\cite{zhang2017convergent}
\begin{align}
  [x]_+ := \max\{0, x\} = \operatorname{prox}_{\imath_{\ge 0}}(x) = \arg\min\nolimits_{z\ge 0} \norm{z-x}^2/2,
\end{align}
where $\operatorname{prox}_f(x) := \arg\min_z f(z) + \norm{z-x}^2/2$ is called
the proximal mapping for a convex function $f$. Hence, the feed-forward
computations in ReLU networks can be approximated by determining the
minimizer $z^*$ of the following convex objective,
\begin{align}
  E(z; x) &:= \tfrac{1}{2} \norm{z_1 - W_0 x}^2 + \tfrac{1}{2} \sum\nolimits_{k=1}^{L-1} \gamma^k \norm{z_{k+1} - W_k z_k}^2,
  \label{eq:free_energy}
\end{align}
subject to $z_k \in \sC_k\subseteq \mathbb R^{n_k}$. $x$ is the input to the
network, $W_k$ are the weights connecting layers $k$ and $k+1$, the parameter $\gamma>0$ is the
feedback weight, and $\sC_k$ are convex sets in $\mathbb R^{n_k}$, where $n_k$
is the dimension of $z_k$. If $\sC_k = \mathbb R^{n_k}$, then the network has linear
units, and if $\sC_k = \mathbb R_{\ge 0}^{n_k}$, then one obtains ReLU
non-linearities. For notational simplicity we omit the bias terms but it can easily be incorporated into the convex formulation. For the output layer $L$ we will always assume $C_L = \mathbb
R^{n_L}$. We call $E(z;x)$ in Eq.~\ref{eq:free_energy} the \emph{free} energy,
since the output activations are not influenced by a target label.

First order optimality conditions on the activations $z_k^*$ read as
\begin{align}
  (\mI + \gamma W_k^T W_k) z_k^* - W_{k-1} z_{k-1}^* - \gamma W_k^T z_{k+1}^* + \partial\imath_{\sC_k}(z_k^*) = 0.
\end{align}
For linear units the subgradient $\partial\imath_{\sC_k}(z_k^*)$ is the zero
vector, and for ReLU units ($\sC_k = \mathbb R_{\ge 0}^{n_k}$) the subgradient
$\partial\imath_{\sC_k}(z_k^*)$ is the non-positive orthant. For small feedback
weights $\gamma\approx 0$ we have $(\mI + \gamma W_k^T W_k) \approx (\mI -
\gamma W_k^T W_k)^{-1}$ and therefore
\begin{align}
  z_k^* \approx \Pi_{\sC_k}\left( W_{k-1} z_{k-1}^* + \gamma W_k^T z_{k+1}^* \right) \stackrel{\gamma\to 0^+}\to \Pi_{\sC_k}\left( W_{k-1} z_{k-1}^* \right).
\end{align}
Thus, by letting the feedback weight $\gamma$ approaching~0 one can emulate
the standard feedforward computation with a lifted network.

\paragraph{Learning with lifted networks}
If a training set $\{(x_i,y_i)\}_{i=1}^N$ is given, then learning with lifted
networks is performed by jointly minimizing over weights and activations~\cite{taylor2016training,zhang2017convergent,askari2018lifted},
\begin{align}
  J_0(\mW) = \frac{1}{N} \sum\nolimits_i \min_{z} \big( E(z) + \ell(z_L, y_i) \big),
\end{align}
where $\ell(\cdot,\cdot)$ is a task-specific loss function. Thus, the aim is
to determine weight matrices $\{W_k\}$ such that the loss-augmented network
energy is as small is possible. Due to the min-min structure of the objective,
minimization by coordinate or block-coordinate descent is possible.
Let us assume $\ell(z_L,y) = \imath\{z_L=y\}$. $J_0$ is then
\begin{align}
  J_0(\mW) = \frac{1}{N} \sum_i \min_{z:z_L=y_i} E(z) = \frac{1}{N} \sum_i \min_z \hat E(z),
  \label{eq:lifted_loss}
\end{align}
where we introduced the \emph{clamped} energy,
\begin{align}
  \hat E(z; x, y) &:= E(z; x) + \imath\{z_L = y\} \nonumber \\
  &= \tfrac{1}{2} \norm{z_1-W_0 x}^2 + \sum\nolimits_{k=1}^{L-1} \tfrac{\gamma^k}{2} \norm{z_{k+1} - W_k z_k}^2 + \tfrac{\gamma^{L-1}}{2} \Norm{y-W_{L-1}z_{L-1}}^2
  \label{eq:clamped_energy}
\end{align}
(subject to $z_k \in \sC_k$), in which input and output units are fixed
(clamped) to given values. Due to the quadratic term
$\Norm{y-W_{L-1}z_{L-1}}^2$, Eq.~\ref{eq:clamped_energy} corresponds to
learning with a quadratic loss. For a given training sample $(x,y)$ and
inferred activations $\hat z$ (such that
$\hat z = \arg\min \hat E(z; x, y)$) the weights are updated such that
pre-activations $W_k \hat z_{k-1}$ and post-activations $\hat z_k$ match. More
precisely, one has (for a single training sample $(x,y)$)
\begin{align}
  \nabla_{W_k} \hat E(\hat z; x, y) = 0 &\iff (W_k \hat z_k - \hat z_{k+1}) \hat z_k^T = 0 \nonumber \\
  &\iff \forall j,j': \hat z_{kj} = 0 \;\lor\; (W_k \hat z_k)_{j'} - \hat z_{k+1,j'} = 0.
\end{align}
This means that $z_{k,j}=0$ whenever there exists $j'$ such that $(W_k \hat
z_k)_{j'} \ne \hat z_{k+1,j'}$ (i.e.\ the network layer behaves non-linearly).
The condition $\nabla_{W_k} \hat E=0$ implies that the layer behaves linearly for
non-vanishing layer inputs $\hat z_k$. Curiously, this is true for non-linearities other than the ReLU.

For a complete training dataset we can deduce that lifted networks trained via Eq.~\ref{eq:lifted_loss} have a strong
tendency to yield linear networks even when trained with constrained (and therefore
non-linearly behaving) activations. This explains the limited accuracies
reported in the literature for such lifted networks, but in a sense this also
justifies to use them for
pre-training~\cite{zhang2017convergent,askari2018lifted}, as they are less
susceptible to vanishing gradients due to their preference for linear
behavior. We validate these claims experimentally in Section~\ref{sec:results}.

\section{Contrastive Learning and Lifted Networks}

In this section, for a single training sample $(x,y)$ we consider a contrastive
variant of Eq.~\ref{eq:lifted_loss},
\begin{align}
  J_1(\mW; x,y) &= \min\nolimits_z \hat E(z) - \min\nolimits_z E(z) = \min\nolimits_{z: z_L = y} E(z) - \min\nolimits_z E(z),
  \label{eq:contrastive_loss}
\end{align}
which we term \emph{contrastive loss}. For a complete training set the
contrastive loss is the average of individual contrastive losses. We call
$\hat z := \arg\min_z \hat E(z)$ the \emph{clamped solution} and $\check z :=
\arg\min_z E(z)$ the \emph{free solution}. By construction $J_1$ is always
non-negative (since $\hat E$ adds the constraint $z_L=y$ to $E$). $J_1$ is a
min-max (``adversarial'') loss,
\begin{align}
  J_1(\mW) &= \min\nolimits_{\hat z} \hat E(\hat z) + \max\nolimits_{\check z} -E(\check z)
  = \min\nolimits_{\hat z} \max\nolimits_{\check z} \left\{ \hat E(\hat z) - E(\check z) \right\}.
\end{align}
Hence, optimization of $J_1$ by alternating minimization of $\mW$, $\hat z$
and $\check z$ is not possible. Using convex duality, one can replace
$\min_{\check z} E(\check z)$ with $\max_\lambda E^*(\lambda)$, yielding
\begin{align}
  J_1(\mW) &= \min\nolimits_{\hat z} \hat E(\hat z) - \min\nolimits_{\check z} E(\check z)
  = \min\nolimits_{\hat z} \hat E(\hat z) - \max\nolimits_\lambda E^*(\lambda)
  = \min\nolimits_{\hat z, \lambda} \left\{ E(\hat z) - E^*(\lambda) \right\}.
\end{align}
Recall that $E^*$ is concave and therefore $-E^*$ convex. For the usual choice
of $\sC_k$ constraint qualification holds and therefore one has strong
duality, $\min_{\check z} E(\check z)=\max_\lambda E^*(\lambda)$. Below we
state the duals programs for $\hat E$ and $E$.

\paragraph{Dual programs}
Via Fenchel or Lagrange duality it can be shown that the dual programs
corresponding to the free and clamped energy, respectively, are given by
\begin{align}
  E^*(\lambda) &= -\sum\nolimits_{k=1}^{L-1} \gamma^{k-1} \left( \frac{\norm{\lambda_k}^2}{2} + \phi_k^*(\gamma W_k^T \lambda_{k+1} - \lambda_k) \right) - \lambda_1^T W_0 x - \imath\{\lambda_L=0\} \\
  \hat E^*(\lambda) &= -\sum\nolimits_{k=1}^{L} \frac{\gamma^{k-1}}{2} \norm{\lambda_k}^2
                      - \sum\nolimits_{k=1}^{L-1} \gamma^{k-1}\phi_k^*(\gamma W_k^T\lambda_{k+1} - \lambda_k) - \lambda_1^T W_0 x + \gamma^{L-1} \lambda_L^T y,
\end{align}
where $\phi_k^*$ is the convex conjugate of $\imath_{\sC_k}(\cdot)$. For
linear units ($\sC_k=\mathbb{R}^{n_k}$) we have $\phi_k^*=\imath_{\{0\}}$
(corresponding to equality constraints), for ReLU units
($\sC_k=\mathbb{R}^{n_k}$) we obtain $\phi_k^* = \imath_{\le 0}$ (i.e.\
inequality constraints), and for the hard sigmoid non-linearity $\sC_k =
[0,1]^{n_k}$ we obtain an $L^1$-like penalizer, $\phi_k^*(\cdot) =
\norm{[\lambda_k - \gamma W_k^T \lambda_{k+1}]_+}_1$. The important relation
used in the following is the connection between optimal primal and dual
variables,
\begin{align}
  \lambda_k = z_k - a_k = z_k - W_{k-1} z_{k-1} ,
\end{align}
which holds for the dual free and clamped energy. $a_k := W_{k-1} z_{k-1}$ is
the pre-activation, i.e.\ the unconstrained signal propagated from layer $k-1$.

\paragraph{A connection between the contrastive loss and learning the posterior}
By observing that optimization over unknowns corresponds approximately to
marginalization (via log-sum-exp), one can restate
Eq.~\ref{eq:contrastive_loss} as follows,
\begin{align}
  J_1(\mW; x, y) = -\lim_{\tau \to 0} \frac{1}{\tau} \log\left( \sum\nolimits_z e^{-\tau E(z; x, y,\mW)} \right) - \frac{1}{\tau} \log\left( \sum\nolimits_{y',z} e^{-\tau E(z; x, y',\mW)} \right),
\end{align}
where $(x,y)$ is a training sample and $\mW$ are the weights made explicit.
%
Let $p(x,y,z; \mW,\tau) = \exp(-\tau E(z; x,y, \mW))/Z(\mW,\tau)$ be the induced
joint probability (at inverse temperature $\tau>0$), then marginalization yields
\begin{align}
  p(x, y; \mW,\tau) &= \frac{1}{Z(\mW,\tau)} \sum\nolimits_z e^{-\tau E(x, y, z, \mW)} & & \text{and} &
  p(x; \mW,\tau) &= \frac{1}{Z(\mW,\tau)} \sum\nolimits_{y',z} e^{-\tau E(x, y', z,\mW)} \nonumber,
\end{align}
and $J_1$ can be restated as
\begin{align}
  J_1(\mW; x, y) &= -\lim_{\tau \to 0} \frac{1}{\tau} \log \frac{p(x, y; \mW,\tau)}{p(x; \mW)} = -\lim_{\tau \to 0} \frac{1}{\tau} \log p(y | x; \mW,\tau).
\end{align}
This means that the contrastive loss (i.e.\ energy of clamped solution minus
energy of free solution) is essentially maximizing the posterior of the
(given) output conditioned on the input. It also means that the free energy
can be interpreted as the unnormalized negative log-likelihood of the input
$x$. Our initial experiments indicate, that $\min_z E(z; x)$ (with weights
obtained by contrastive training) is only of limited use to directly score
inputs $x$, i.e.\ to assign log-likelihoods. Due to the shape of the free
energy Eq.~\ref{eq:free_energy} only the first layer activations contribute to
$E(z;x)$ in the weak feedback setting. Nevertheless, it opens a possible new
perspective of understanding DNNs.

\section{Contrastive Learning Approximates Back-propagation}
\label{sec:CHL_and_BP}

Contrastive learning of weights by minimizing $J_1(\mW)$ approximates
back-propagation for small values of $\gamma$. This was shown for a particular
non-convex free energy (related to contrastive Hebbian learning~\cite{movellan1991contrastive})
in~\cite{xie2003equivalence}. \cite{scellier2017equilibrium} establishes
establishes a general connection between nested optimization problems and
contrastive objectives. In this framework the contrastive cost $J_1(\mW)$ is
interpreted as a finite difference approximation,
\begin{align}
  \tfrac{1}{\beta} J_1(\mW) &= \tfrac{1}{\beta} \left( \min_z \left\{ E(z;x,\mW) + \beta \norm{z_L - y}^2 \right\} - \min_z E(z;x,\mW) \right)
\end{align}
evaluated at $\beta=0$. In our formulation $\beta = \gamma^{L-1}$. Letting
$\beta\to 0$ (i.e.\ $\gamma\to 0$) it is shown that
\begin{align}
  \lim_{\beta\to 0} \tfrac{1}{\beta} \nabla_{\mW} J_1(\mW) = \nabla_{\mW} \norm{z_L^* - y}^2 \qquad \text{s.t. } z^* = \arg\min_z E(z; \mW),
\end{align}
i.e.\ the gradient of the contrastive loss approaches the gradient of a nested
optimization problem. Since for $\gamma\to 0$, $z^* = \arg\min_z E(z; \mW)$
converges to the activations obtained by a standard forward pass, one can
conclude that $\frac{1}{\gamma^{L-1}} \nabla_{\mW} J_1(\mW) \stackrel{\gamma\to
  0}\to \nabla_\mW \norm{z_L^* - y}^2/2$.\footnote{A further small
  complication is that $E$ is required to be differentiable in
  $z$. This can be achieved by replacing e.g.\ non-negativity constraints on $z$
  by penalizers or barriers such as $-\eps\log(z_k)$ for $\eps \approx 0$.}

The main shortcoming of the above ``algebraic'' and indirect derivation is
that the structural properties of $E$ are largely ignored, and that for
finite (non-infinitesimal) values of $\gamma$ it is not clear how contrastive
learning deviates from back-propagation. Hence, in the following we provide a
direct and constructive proof.

By recalling the connection between primal and dual variables ($\lambda_k =
z_k - a_k$, see Section~\ref{sec:contrastive_loss}) we can write the gradient
of $J_1$ w.r.t.\ $W_k$ concisely as follows:
\begin{align}
  \nabla_{W_k} J_1(\mW) &= \nabla_{W_k} \left( E(\hat z; \mW) - E(\check z; \mW) \right)
  = \gamma^{k} (W_k \hat z_k - \hat z_{k+1}) \hat z_k^T - \gamma^{k-1} (W_k \check z_k - \check z_{k+1}) \check z_k^T \nonumber\\
  &= -\gamma^{k} \left( \hat\lambda_{k+1} \hat z_k^T - \check\lambda_{k+1} \check z_k^T \right).
  \label{eq:gradient_J}
\end{align}

\paragraph{Weight updates for linear networks}
Let us focus on $C_k=\mathbb R^{n_k}$, i.e.\ network activations $z_k$ are
unconstrained. First note that in this setting the free phase has zero cost,
$\min_z E(z) = 0$. Hence, $J_1(\mW)$ reduces to $\min_z \hat E(z;\mW)$ and
\begin{align}
  \nabla_{W_k} J_1(\mW) = -\gamma^{k} \lambda_{k+1} \hat z_k^T.
  \label{eq:grad_linear_units}
\end{align}
From $C_k=\mathbb R^{n_k}$ we deduce that $\phi_k^* =
\imath_0$ and therefore the dual clamped energy reads as
\begin{align}
  \hat E^*(\lambda) &= -\sum\nolimits_{k=1}^{L} \frac{\gamma^{k-1}}{2} \norm{\lambda_k}^2 - \lambda_1^T W_0 x + \gamma^{L-1} \lambda_L^T y,
                      \label{eq:dual_E_linear}
\end{align}
subject to $\lambda_k = \gamma W_k^T \lambda_{k+1}$. By recursively expanding
this constraint we can express the dual variables solely in terms of $\lambda_L$,
\begin{align}
  \lambda_k = \gamma^{L-k} \prod\nolimits_{j=k}^{L-1} W_j^T \lambda_L.
  \label{eq:lambda_expanded_linear}
\end{align}
Since $\lambda_L = \hat z_L - \hat a_L = y - W_{L-1} \hat z_{L-1}$, i.e.\ the difference between target
and predicted output, we introduce $\Delta y := \lambda_L = y - W_{L-1}
\hat z_{L-1}$. Consequently, Eq.~\ref{eq:grad_linear_units} can be restated as
\begin{align}
  \label{eq:expanded_grad_linear_units}
  \nabla_{W_k} J_1(\mW) &= -\gamma^{k} \gamma^{L-k-1} \prod\nolimits_{j=k+1}^{L-1} W_j^T \Delta y\, \hat z_k^T
  = \gamma^{L-1} \prod\nolimits_{j=k+1}^{L-1} W_j^T (W_{L-1} \hat z_{L-1} - y) \hat z_k^T.
\end{align}
The error signal arriving at layer $k$ is $\Delta y^{(k)} :=
\prod_{j=k+1}^{L-1} W_j^T (W_{L-1} \hat z_{L-1} - y)$, which is the same as
the error signal used in back-propagation. If we denote the forward propagated
value by $x^{(k)} := W_{k-1}\cdots W_0 x$, then the only difference (besides
the constant scaling $\gamma^{L-1}$) to the gradient induced by
back-propagation, $\nabla_{W_k}^{\text{back-prop}} = \Delta y^{(k)} (x^{(k)})^T$,
is the occurrence of $\hat z_k$ instead of $x^{(k)}$. In the weak feedback
setting ($\gamma \ll 1$) we have $\hat z_k \approx x^{(k)}$.  Thus, for finite
(non-infinitesimal) values of $\gamma$, the difference between
back-propagation induced parameter updates and the ones given by contrastive
learning lies in the difference of the utilized activations (pure forward vs.\
inferred). Further, in light of Eq.~\ref{eq:lambda_expanded_linear} we also
expect the contrastive loss-based gradients for earlier layers to be closer to
back-propagation gradients than later layers.

\paragraph{Weight updates for ReLU networks}
If we add non-negativity constraints on the hiddens, i.e.\ $C_k = \mathbb
R_{\ge 0}^{n_k}$, then $\phi_k^* = \imath_{\le 0}$ and the dual objective is
given by Eq.~\ref{eq:dual_E_linear},
but with different constraints,
$\lambda_k \ge \gamma W_k^T \lambda_{k+1}$. These constraints can be stated as
$\lambda_k = \gamma W_k^T \lambda_{k+1} + \nu_k$ for $\nu_k \ge 0$. By
inserting the connection between primal and dual variables,
$\lambda_k = z_k - a_k$, we obtain
\begin{align}
  z_k - W_{k-1} z_{k-1} = \gamma W_k^T (z_{k+1} - W_k z_k) + \nu_k \qquad\text{or} \nonumber \\
  (\m I + \gamma W_k^T W_k) z_k - W_{k-1} z_{k-1} - \gamma W_k^T z_{k+1} - \nu_k = 0,
\end{align}
which can be identified as first order optimality condition for $z_k$. Hence,
$-\nu_k \in \partial \imath_{\ge 0}(z_k)$, and we also have complementary
slackness: $z_{kj}>0$ implies $\nu_{kj} = 0$, and if the constraint $z_{kj}\ge 0$ is
active, then $\nu_{kj} > 0$.

We group the activations $z_k$ into strictly positive elements and clamped
(i.e.\ zero) ones. After permuting indices, such that strictly positive
activations come first, one can write $z_k = (\tilde z_k^T,\;\v 0^T)^T$,
where $\tilde z_k > 0$ corresponds to elements in $z_k$ where the
non-negativity constraints are inactive. Hence, the relation between primal
and dual variables is
\begin{align}
  \begin{pmatrix} \tilde \lambda_k \\ \lambda_k^0 \end{pmatrix} = \begin{pmatrix} \tilde z_k \\ \v 0 \end{pmatrix}
  - \begin{pmatrix} \tilde a_k \\ a_k^0 \end{pmatrix},
\end{align}
where $a_k^0$ and $\lambda_k^0$ correspond to clamped component in
$z_k$. Since $\tilde z_k > 0$, the corresponding dual variable $\tilde\nu_k$
are zero (via complementary slackness), and we obtain $\tilde\lambda_k = \gamma \tilde W_k^T \lambda_{k+1}$,
We recall Eq.~\ref{eq:gradient_J},
\begin{align}
  \nabla_{W_k} J_1(\mW) &= -\gamma^{k} \left( \hat\lambda_{k+1} \hat z_k^T - \check\lambda_{k+1} \check z_k^T \right)
  \approx -\gamma^{k} (\hat\lambda_{k+1} - \check\lambda_{k+1}) \hat z_k^T,
  \label{eq:gradient_J_ReLU}
\end{align}
where we assume weak feedback ($\gamma\ll 1$), and therefore the activations
of the free and clamped phase are close, i.e.\ $\check z_k \approx \hat z_k$
and $\check z_{k+1} \approx \hat z_{k+1}$. This assumption also implies that
the free and clamped solutions share the set of clamped activations. By
recursively expanding the relation $\lambda_{k+1} = \gamma W_{k+1}^T
\lambda_{k+2} + \nu_{k+1}$ we therefore obtain
\begin{align}
  \hat\lambda_{k+1} - \check\lambda_{k+1} &= \gamma\begin{pmatrix} \tilde W_{k+1}^T (\hat\lambda_{k+2}-\check\lambda_{k+2}) \\ \v 0\end{pmatrix}
  = \gamma \operatorname{diag}(\hat z_{k+1} > 0) W_{k+1}^T (\hat\lambda_{k+2} - \check\lambda_{k+2}) \nonumber \\
  &= \gamma^2 \operatorname{diag}(\hat z_{k+1} > 0) W_{k+1}^T \operatorname{diag}(\hat z_{k+2} > 0) W_{k+2}^T (\hat\lambda_{k+3} - \check\lambda_{k+3}) \nonumber \\
  &= \gamma^{L-k-1} \left( \prod\nolimits_{k'=k+1}^{L-1} \operatorname{diag}(\hat z_{k'}>0) W_{k'}^T \right) \underbrace{(\hat\lambda_L-\check\lambda_L)}_{=\Delta y}.
\end{align}
This equation is almost exactly the error signal used in back-propagation (the
difference being that $\hat z_k$ is appearing instead of the purely forward
propagated $x^{(k)}$). This implies that the contrastive loss leads
(approximately) to the correct cancellation of error signals propagated
backwards. Inserting this into Eq.~\ref{eq:gradient_J_ReLU} yields
\begin{align}
  \nabla_{W_k} J_1(\mW) &\approx -\gamma^k (\hat\lambda_{k+1}-\check\lambda_{k+1}) \hat z_k^T
  = -\gamma^k \gamma^{L-k-1} \left( \prod\nolimits_{k'=k+1}^{L-1} \operatorname{diag}(\hat z_{k'}>0) W_{k'}^T \right) (\hat\lambda_L-\check\lambda_L) \hat z_k^T
  \nonumber \\
  &= \label{eq:expanded_gradient_J_ReLU}
    -\gamma^{L-1} \left( \prod\nolimits_{k'=k+1}^{L-1} \operatorname{diag}(\hat z_{k'}>0) W_{k'}^T \right) (\hat\lambda_L-\check\lambda_L) \hat z_k^T \\
  &\approx -\gamma^{L-1} \left( \prod\nolimits_{k'=k+1}^{L-1} \operatorname{diag}(\hat z_{k'} > 0) W_{k'}^T \right) \Delta y\,(x^{(k)})^T,
\end{align}
where the last line corresponds to standard back-propagation. We can summarize
this section as follows: in contrast to regular losses for lifted networks the
use of the contrastive loss yields (approximately) the correct backward signal
and weight update. Standard lifted networks do not approximate
back-propagation even in the small feedback setting.

\begin{remark}
  The analogous result can be obtained if non-negativity constraints are
  replaced by more general element-wise bounds constraints, such as $z_k \in
  [0,1]^{n_k}$.
\end{remark}

\section{Numerical Validation}
\label{sec:results}

The aim of this section is to numerically verify that (i) standard training
for lifted networks has a strong bias towards linear behavior, and (ii) that
contrastive training of lifted networks yields results comparable to
back-propagation.

\paragraph{Implementation}
We use a straightforward C++ implementation with multi-threading
acceleration. Inference in lifted networks requires solving a convex quadratic
program subject to optional bound constraints. We use a coordinate descent
method that traverses the layers and updates a single element $z_{kj}$ in each
step. Updating $z_{kj}$ can be done in closed form, and each activation
$z_{kj}$ is updated 15 times. Since we use weak feedback ($\gamma=1/8$ in our
experiments), we initialize $z$ via a regular forward pass (i.e.\
$z_{k+1} = \Pi_{\sC_{k+1}}(W_k z_k)$ with $z_0=x$). For this choice of
$\gamma$ the classification accuracies obtained by pure forward passes and by
minimization of activations are almost identical.

ReLU networks are obtained by setting
$\sC_k = \mathbb{R}_{\ge 0}^{n_k}$, and hard sigmoid non-linearities use
$\sC_k = [0,1]^{n_k}$. We also include results for linear regression to
further support our claim that standard, non-contrastively trained lifted
networks essentially behave like linear regressors.
Although we omitted bias terms in
the equations, they are used in our implementation. For training the weight
matrices are initialized element-wise with random values from a normal
distribution, and biases are initialized to~0. The different training
approaches start from the same initial network weights.
After the clamped and free activations are determined, the weights are updated
using stochastic gradient descent (with mini-batches of size~50). In view of
Eqs.~\ref{eq:expanded_grad_linear_units} and~\ref{eq:expanded_gradient_J_ReLU}
suitable learning rates $\eta^{\mathrm{BP}}$ for back-propagation and
$\eta^{\mathrm{Contr.}}$ for contrastive learning are approximately related by
$\eta^{\mathrm{Contr.}} = \eta^{\mathrm{BP}}/\gamma^{L-1}$.
We use constant learning rates $\eta^{\mathrm{BP}} = 1/20$ for
back-propagation and contrastive learning. Standard training of lifted
networks is more difficult: in order to at least match the performance of
linear regression, the first epochs utilized a smaller learning rate.
We use 100 epochs for the MNIST and Fashion-MNIST datasets, and 50 epochs for
the CIFAR-10 dataset.

\begin{figure}[htb]
  \centering
  \subfigure[Back-prop (99\%/97.5\%)]{\includegraphics[width=0.27\textwidth]{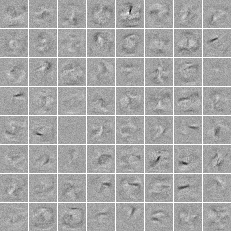}}
  \subfigure[Contr.\ lifted (99\%/97.3\%)]{\includegraphics[width=0.27\textwidth]{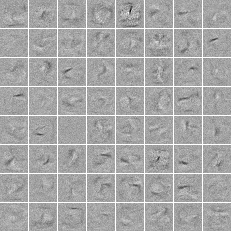}}
  \subfigure[Standard lifted (85\%/86\%)]{\includegraphics[width=0.27\textwidth]{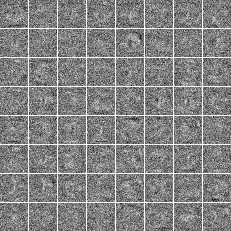}}
  \caption{First layer filters of a 784-64-64-10 ReLU network trained on
    MNIST. Training and test accuracies are given in parentheses.}
  \label{fig:first_layer_mnist}
\end{figure}

\begin{table}[htb]
  \centering
  \begin{tabular}{|l|c c|c c c|}
    \hline 
     & \multicolumn{2}{|c|}{accuracy (\%)} & \multicolumn{3}{|c|}{linear activations (\%)}  \\ 
    models & train & test  & layer 1 & layer 2 & layer 3 \\ 
    \hline 
    ReLU back-prop     & 99.8 & 97.7  & 43.8 & 39.3 & 38.2 \\ 
    ReLU lifted & 85.2 & 86.3  & 99.9 & 99.9 & 99.9 \\
    ReLU contr. & 99.8 & 97.6  & 37.9 & 43.6 & 48.4 \\
    Hard sigm.\ back-prop & 99.7 & 97.0 & 53.5 & 45.1 & 34.6 \\ 
    Hard sigm.\ lifted & 85.4 & 86.3 & 99.9 & 99.9 & 99.9 \\
    Hard sigm.\ contr. & 99.8 & 97.4 & 47.0 & 43.7 & 50.7 \\
    Linear regression & 85.3 & 86.0  &      &      &      \\
    \hline 
  \end{tabular}
  \caption{Training and test accuracies and fraction of activations in the
    linear range for 784-64-64-64-10 networks trained on MNIST.}
  \label{tab:mnist_comparisons}
\end{table}

\begin{table}[htb]
  \centering
  \begin{tabular}{|l|c c|c c c|}
    \hline 
     & \multicolumn{2}{|c|}{accuracy (\%)} & \multicolumn{3}{|c|}{linear activations (\%)}  \\ 
    models & train & test  & layer 1 & layer 2 & layer 3 \\ 
    \hline 
    ReLU back-prop     & 95.6 & 88.0  & 38.0 & 39.6 & 38.4 \\ 
    ReLU lifted        & 81.4 & 80.0  & 99.9 & 99.9 & 99.9 \\
    ReLU contr.        & 94.2 & 88.2  & 30.0 & 48.2 & 66.1 \\
    Hard sigm.\ back-prop & 95.5 & 88.1 & 46.7 & 40.5 & 41.7 \\ 
    Hard sigm.\ lifted & 81.6 & 80.1 & 99.9 & 99.9 & 99.9 \\
    Hard sigm.\ contr. & 94.7 & 88.2 & 44.9 & 35.6 & 66.3 \\
    Linear regression  & 82.3 & 80.4  &      &      &      \\
    \hline 
  \end{tabular}
  \caption{Training and test accuracies and fraction of activations in the
    linear range for 784-64-64-64-10 networks trained on Fashion-MNIST.}
  \label{tab:fashion_mnist_comparisons}
\end{table}

\paragraph{MNIST and Fashion-MNIST}
In Fig.~\ref{fig:first_layer_mnist} the first layer weights of a fully
connected 784-64-64-10 ReLU network trained on MNIST~\cite{lecun1998gradient}
are visualized. It can be observed that the weights obtained by
back-propagation (Fig.~\ref{fig:first_layer_mnist}(a)) and the ones obtained
by contrastive training for lifted networks
(Fig.~\ref{fig:first_layer_mnist}(b)) are visually close, whereas the weights
returned by standard traininig of lifted networks
(Fig.~\ref{fig:first_layer_mnist}(c)) are visually different. This is also
reflected in the achieved training and test accuracies.
Table~\ref{tab:mnist_comparisons} illustrates results for a 3-layer
784-64-64-64-10 network with either ReLU or hard sigmoid
non-linearities. Back-propagation and contrastive learning achieve again
similar prediction accuracies (substantially better than regular lifted
training). The most interesting aspect in Table~\ref{tab:mnist_comparisons} 
is that contrastive training leads to around 50\% active non-linearities while
regularly trained lifted networks are almost entirely in their linear regime.
Table~\ref{tab:fashion_mnist_comparisons}
and~\ref{tab:fashion_mnist_comparisons2} depict the corresponding results for
the Fashion-MNIST dataset~\cite{xiao2017fashion_mnist} (using four and
five-layer networks, respectively). The results follow the same pattern as the
ones for the standard MNIST dataset.

\begin{table}[htb]
  \centering
  \begin{tabular}{|l|c c|c c c c|}
    \hline 
     & \multicolumn{2}{|c|}{accuracy (\%)} & \multicolumn{4}{|c|}{linear activations (\%)}  \\ 
    models & train & test  & layer 1 & layer 2 & layer 3 & layer 4 \\ 
    \hline 
    ReLU back-prop     & 96.2 & 88.4 & 33.5 & 42.0 & 44.3 & 40.3 \\
    ReLU lifted        & 72.7 & 72.3 & 99.9 & 99.9 & 99.9 & 99.9 \\
    ReLU contr.        & 97.0 & 88.6 & 36.1 & 44.4 & 45.5 & 67.9 \\
    \hline 
  \end{tabular}
  \caption{Training and test accuracies and fraction of activations in the
    linear range for 784-128-64-64-32-10 networks trained on Fashion-MNIST.}
  \label{tab:fashion_mnist_comparisons2}
\end{table}

\paragraph{CIFAR-10}
Table~\ref{tab:cifar_comparisons} illustrates the analogous results for a
grayscale version of the CIFAR-10 dataset~\cite{krizhevsky2009learning}. We
observe qualitatively similar results (although at substantially lower
accuracy levels compared to MNIST and Fashion-MNIST) for this dataset. No
augmentation (such as horizontal flipping of input images) is employed.

\begin{table}[htb]
  \centering
  \begin{tabular}{|l|c c|c c|}
    \hline 
     & \multicolumn{2}{|c|}{accuracy (\%)} & \multicolumn{2}{|c|}{linear activations (\%)}  \\ 
    models & train & test  & layer 1 & layer 2 \\ 
    \hline 
    ReLU back-prop & 71.1 & 48.2  & 28.5 & 28.9 \\ 
    ReLU lifted    & 28.0 & 27.7  & 99.9 & 99.9 \\
    ReLU contr.    & 61.7 & 43.1  & 15.2 & 90.8 \\
    Linear regression & 24.2 & 21.8  &      &     \\
    \hline 
  \end{tabular}
  \caption{Training and test accuracies and fraction of activations in the
    linear range for 1024-512-512-10 ReLU network trained on (grayscale) CIFAR-10. }
  \label{tab:cifar_comparisons}
\end{table}

\section{Conclusion}

The aim of this work is to draw the attention to lifted networks, which---as
easy-to-train energy-based models---are attractive for understanding
DNNs. Lifted networks have seen somewhat limited use in the literature, and we
hypothesize that this is due to the standard training procedures for lifted
networks essentially lead to linear network behavior.  We demonstrate that
using a contrastive training objective leads to more competitive lifted
networks, which also better leverage the expressive power of the network
non-linearities.

In future work we intend to leverage the network energy to estimate the
likelihood of input data, e.g.\ for anomaly detection. We also plan to
investigate early stopping criteria for activation inference, which will allow
faster training procedures for lifted networks.

\small
\bibliographystyle{plain}
\bibliography{literature}

\end{document}